\documentclass[10pt, a4paper]{article}

\usepackage[final]{lrec-coling2024} 

\title{Human and Automatic Interpretation of Romanian Noun Compounds}

\name{Ioana Marinescu, Christiane Fellbaum} 

\address{Department of Computer Science\\
         Princeton University\\
         \{ioanam, fellbaum\}@princeton.edu\\}

\abstract{
    Determining the intended, context-dependent meanings of noun compounds like \emph{shoe sale} and \emph{fire sale} remains a challenge for NLP. Previous work has relied on inventories of semantic relations that capture the different meanings between compound members. Focusing on Romanian compounds, whose morphosyntax differs from that of their English counterparts, we propose a new set of relations and test it with human annotators and a neural net classifier. Results show an alignment of the network's predictions and human judgments, even where the human agreement rate is low. Agreement tracks with the frequency of the selected relations, regardless of structural differences. However, the most frequently selected relation was none of the sixteen labeled semantic relations, indicating the need for a better relation inventory.}

\begin{document}

\maketitleabstract

\Keywords{Romanian noun compounds, semantic roles, human annotations, automatic classification, word embeddings.} 

\section{Introduction}

The interpretation of noun compounds requires construing a meaning relation between the constituent members. Previous work has explicated and classified these relations in terms of finite inventories of semantic roles. Thus, in compounds like \emph{garage sale, shoe sale, fire sale, summer sale, baby sale}, the first member bears a different relation to the head noun \emph{sale} (Location, Object, Circumstance, Time, Beneficiary, respectively). Language users interpret these appropriately:  we sell shoes but not babies, fire, or seasons. Similarly, when \emph{baby} serves as a modifier, as in \emph{baby carrot}, we interpret it not as referring to an infant but to something smaller in size than the norm for its category. Speakers regularly create and interpret novel compounds, making them challenging for programs and infeasible for enumerative listing and automatic “look-up” in dictionaries. 
All endocentric compounds are commonly assumed to be uniquely mappable onto one out of a finite set of labeled relations between a head and its modifier(s), by both humans and automatic systems. We test this assumption, comparing human annotation and the predictions of a neural net for classifying Romanian noun compounds with a novel set of relations.

We focus on two-member compounds, such as \emph{robot arm}, where the relation between \emph{robot}, the modifier, and \emph{arm}, the compound head, can be described as one of part-whole, and  \emph{plastic arm}, where the relation is one of material-whole.
English endocentric noun compounds are composed of two bare nouns, with a head preceded by a modifier. Examples are
\emph{glass bottle,} \emph{ocean water,} \emph{leather jacket,} \emph{morning run},  and \emph{winter cold}. In the Romanian equivalents the head precedes  the modifier, which is marked with the genitive case morphology. An example is \emph{apa oceanului}, (lit. "water ocean-GEN"). In other compounds a preposition joins the two nouns, with the head noun marked for accusative case (\emph{geaca de piele}, lit. "jacket of leather"). Unlike in English, there are no compounds consisting of two bare nouns. 

Our goal is threefold. First, we propose a new set of semantic relations for noun compounds. Second, we evaluate and compare human and automatic mappings of Romanian compounds onto our relations. Third, by considering Romanian compounds, we ask whether their morphosyntax, which differs from English, bears measurably on the their semantic interpretation. Finally, we hope to stimulate further research into the semantic analysis of noun compounds in Romanian and other less studied languages. 


\section{Related work}
Developing a comprehensive inventory of semantic relations has long been a challenge for linguistics 
\cite{nastaseetal2013}.\footnote{It is useful to recall earlier work in linguistics on "semantic" or "thematic" roles, which were intended to capture the meaning relation between a verb and its arguments. Numerous proposals were made, e.g., \cite{Gruber, Fillmore, Dowty} for role inventories at different levels of granularity but no consensus was reached; notably, one role labeled "Theme" covers a wider range of case that seem to elude  specific description.}
Proposals range from a small set of relations (e.g., \citealp{levi}) to a hierarchy of a few primary semantic relations with fine-grained sub-relations \cite{nakov2008} to an unbounded set of relations (\citealp{Downing1977OnTC}). \citet{verhoeven} studied compounds in Dutch and Afrikaans, classifying them in six categories. At present no universally agreed-upon semantic role inventory exists \cite{Kim2007InterpretingNC, tratz-hovy-2010-taxonomy}.

Computational approaches to compound interpretation hold promises for applications such as machine translation and information retrieval  \cite{nakov2008}. 
Automatic classification using machine learning methods with word embeddings such as word2vec \cite{mikolov2013efficient} and GloVe \cite{pennington-etal-2014-glove} include \cite{dima-hinrichs-2015-automatic}.   

 \citet{girju2005} use WordNet \cite{wordnet} and SVMs to distinguish thirty-five semantic relations for noun compounds. \citet{kim-baldwin-2005-automatic} employ a WordNet-based similarity approach to classify nouns into twenty cateogories. \citet{OSEAGHDHA_COPESTAKE_2013} used a SVM approach for a six-way classification. \citet{tratz-hovy-2010-taxonomy} apply a Maximum Entropy classifier for forty-three categories, while \citet{dima-hinrichs-2015-automatic} apply a neural network and word embeddings towards a classification using the same forty-three categories.
 We add to this body of research by investigating Romanian compounds, with a novel set of semantic relations departing from those in  \cite{tratz-hovy-2010-taxonomy}.

\section{Approach}

Our dataset consists of 1000 noun compounds selected from the Romanian Universal Dependency treebank, RoRefTrees \cite{barbu2016romanian}. We propose a novel  taxonomy of semantic relations comprised of sixteen labeled plus one unlabeled categories, modifying the one of  
\cite{tratz-hovy-2010-taxonomy}, and collect human judgments for our compounds via Amazon Mechanical Turk. We implement a machine learning method to evaluate the automatic interpretation of compounds based on BERT word embeddings \cite{devlin2019bert}, concatenating the embeddings for noun constituents. The embeddings constitute the input for a multi-layer perceptron (MLP) that classifies the compounds according to the taxonomy in Table \ref{tab:taxonomy}.

\subsection{Dataset}
We extract noun compounds from the Romanian Universal Dependency treebank \cite{barbu2016romanian}, a balanced corpus including texts from diverse genres. 
Out of the 185,113 tokens in the corpus, 45,988 are nouns. We extract sequences of two nouns with and without an intervening preposition (e.g. \emph{apa oceanului} and \emph{geaca de piele}) and manually check for false positives. 

For each word that occurs at least once in the extracted compounds, we count (i) the overall  frequency, (ii) the number of compounds where that word is the compound head, and (iii) the number of compounds where it is the modifier. 
We select 1100 compounds based on the frequency of the head noun, one compound for each head noun type. We manually reduce this set to a total of 1000 compounds to eliminate false positives (non-compounds). While frequency and polysemy are known to be positively correlated, we assume that within a compound, the nouns will have just a single sense.

The majority of the compounds (n=5547) show the NPN syntax. Among these, roughly half (n=2866) contain the preposition \emph{de}, as in \emph{unitate de time} ("unit of time"). 
Our data include 3370 NN compounds, where the second noun (the modifier) is marked with the genitive case (\emph{frigul iernii}, "winter cold;" \emph{timpul noptii}, "night time"). 
Note that, unlike in English, two consecutive nouns without case marking do not constitute a compound in Romanian.

\subsection{A novel semantic relation taxonomy}

The challenge for compound interpretation is to choose a set of labeled semantic relations that each capture the way in which meanings of the compound members interact to form a compound meaning. The relations should be unambiguous and discrete, i.e., non-overlapping. While this might invite a small set of relatively coarse-grained relations, these may overlook the kind of subtler semantic distinctions that humans easily draw. On the other hand, with too large an inventory of relations, discrimininability might decrease and human annotators creating a gold standard might have trouble learning and applying the distinctions. 

We depart from the forty-three relations proposed by \cite{tratz-hovy-2010-taxonomy} and propose a reduced set, on the reasoning that learning and distinguishing a large number of categories is difficult and may lead to errors. Moreover, such complexity would be exacerbated for deep learning systems. To ensure that the task is manageable for both humans and machine learning, we restrict our classification to sixteen categories labeled categories, plus one for all cases not covered by these sixteen and labeled "none," meaning "neither of the listed choices." Table \ref{tab:taxonomy} lists the relations and provides examples. The category names follow the Romanian order of head and modifier, the reverse of the English order.  

We modify \cite{tratz-hovy-2010-taxonomy}'s taxonomy, where fine-grained categories like "communicator of communication" and "performer of activity" are grouped into broader, basic semantic relations such as "causal." From each of their broader groups, we adopt some sub-relations but alter the labels such that they more clearly express the nouns' semantic roles. For example, from the broad "Purpose/Activity" relation, we adopt the category "modify/process/change" but rename it "process + undergoer." This category includes \emph{eye surgery} in \citet{tratz-hovy-2010-taxonomy} and accommodates the Romanian equivalent \emph{operatie la ochi}. We do not adopt groups that we deemed to broad or vague, such as "Topic." \citet{tratz-hovy-2010-taxonomy}'s category "Other" corresponds to our "none," and is illustrated by \emph{lentile de contact (contact lens)}.

\begin{table}[h]
\centering
\resizebox{\columnwidth}{!}{%
\begin{tabular}{|c|l|l|}
\hline
\textbf{ID} & \textbf{Category Name}                           & \multicolumn{1}{c|}{\textbf{Examples}}                                            \\ \hline
1           & None of the categories                          & \textit{lentile de contact (contact lens), efectul fluturelui (butterfly effect)} \\ 
2           & Process + undergoer                             & \textit{operatie la ochi (eye surgery), modificarea legii (law change)}           \\ 
3           & Entity + scope                                  & \textit{tren de marfa (cargo train), sac de dormit (sleeping bag)}                \\ 
4           & Entity + attribute                              & \textit{culoarea pamantului (earth color), poveste de dragoste (love story)}      \\ 
5           & Result + cause                                  & \textit{castigurile din reclame (ads revenue), daunele inundatiei (flood damage)}  \\ 
6           & Event + agent                                   & \textit{abuzul politiei (police abuse), plansul copilului (child's cry)}           \\ 
7           & Possession + possessor                          & \textit{averea familiei (family fortune), bratara fetei (girl's bracelet)}         \\ 
8           & Entity/process/result + cause/source           & \textit{deficienta de vitamine (vitamin deficit), declaratie de razboi (war declaration)} \\ 
9           & Detachable part + whole                        & \textit{bratul robotului (robot arm), piciorul scaunului (chair leg)}              \\ 
10          & Tool + operation/undergoer                     & \textit{deschizator de conserve (can opener), perie de lustruit (polishing brush)} \\ 
11          & Location + locatum                             & \textit{aragazul din bucatarie (kitchen stove), magazin de pantofi (shoe store)}    \\ 
12          & Time + event                                    & \textit{data nasterii (birth date), perioada examenului (exam period)}              \\ 
13          & Experience/emotion + experiencer               & \textit{anxietatea studentului (student anxiety), admiratia fanilor (fan admiration)} \\ 
14          & Substance/material part + whole                & \textit{supa de mazare (pea soup), punga de plastic (plastic bag)}                 \\ 
15          & Event + time of event                          & \textit{tura de noapte (night shift), alergat de dimineata (morning run)}           \\ 
16          & Beneficit + beneficiary                        & \textit{mancare de pisica (cat food), beneficii de somer (unemployment benefits)}   \\ 
17          & Duration + event                                & \textit{intalnire de o ora (hour meeting), excursie de o zi (day trip)}            \\ \hline
\end{tabular}%
}
\caption{Semantic classification}
\label{tab:taxonomy}
\end{table}

\subsection{Collecting human judgments}
We collected annotations using Amazon Mechanical Turk. Each of our 1000 compounds received an annotation from two of our nine Romanian native speakers.  Prior to annotating, they were shown two example compounds for each category accompanied by an explanation. 
For the actual task, the  annotators saw a screen with a window displaying a compound and the seventeen labeled options, and they were asked to select one with a mouse click.  We discarded the judgments of annotators who labeled less than twenty compounds.

\subsection{Model and evaluation}
\label{model architecture}
We evaluate the automatic interpretation of the compounds. Our model uses the Romanian BERT embeddings \cite{dumitrescu-etal-2020-birth}, which give a 768-dimensional representation of the two nouns in the compounds. We split our dataset into a training set (750 compounds) and test set (250 compounds). We trained a simple 2-layer multi-layer perceptron that takes as input the concatenated embeddings for the two nouns in the compound. The training labels are either a one-hot vector if the annotators agreed on the classification, or, if the two annotators each chose a different category, a probability of 0.5 for each of the categories selected. We trained the model using Stochastic Gradient Descent (SGD), learning rate 0.01, and Cross-Entropy loss.

We evaluate the model as follows: when the annotators' judgments agreed, we choose the answer to be the category with the highest probability from the output of our model; when the annotators disagreed, we looked at the two categories with the highest probabilities and checked if there is an overlap between these and the two categories proposed by the human annotators. 

\section{Results}
We discuss the results of the human and the automatic classification of the compounds. 

\begin{figure}
    \centering
    \includegraphics[width=0.6\linewidth]{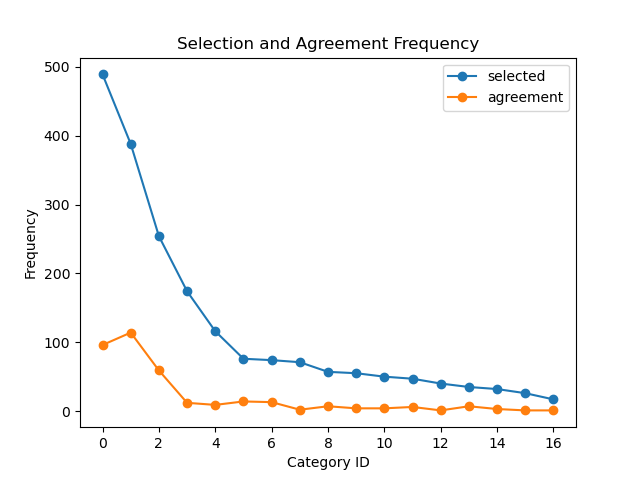}
    \caption{Frequency of selection and frequency of agreement. The most frequently selected categories show the highest inter-annotator agreement.}
    \label{fig:agreement_freq}
\end{figure}

\textbf{Human annotations.} Figure \ref{fig:agreement_freq} shows how many times each of the categories was selected (out of $2 \times 1000$ possible selections) and how many times the category was selected by both annotators for a given compound.\footnote{The distribution of total number of compounds labeled by each annotator is skewed so we could not compute Cohen's $\kappa$.} Out of 1000 noun compounds, 352 received the same label from both annotators. Figure \ref{fig:agreement_freq} shows that the most frequently selected categories are the ones that annotators agree most on. 


\begin{figure}[htbp]
    \begin{minipage}{0.5\linewidth}
        \centering
        \includegraphics[width=\linewidth]{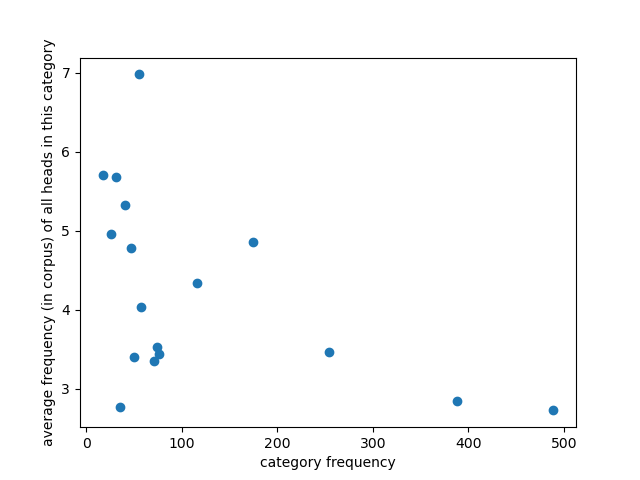}
    \end{minipage}%
    \begin{minipage}{0.5\linewidth}
        \centering
        \includegraphics[width=\linewidth]{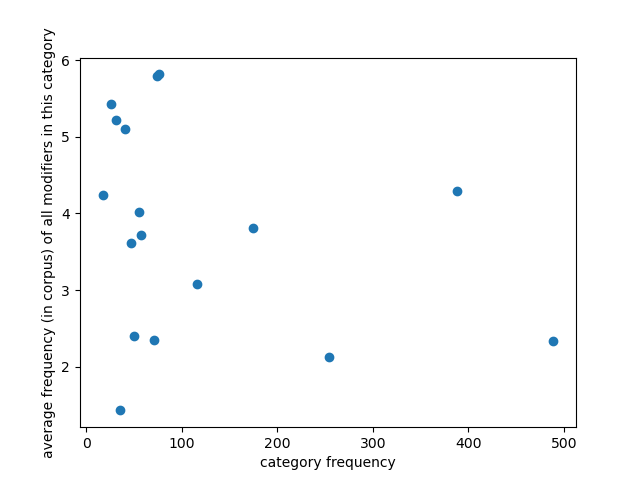}
    \end{minipage}
    \caption{Average frequency (in all noun compounds in Romanian UD treebank) of the heads and modifiers, respectively, of the compounds in each category as a function of how many times that category was selected by the annotators}
    \label{fig:freqs_category_heads_modifiers}
\end{figure}

\citet{butnariu-etal-2010-semeval} found that noun compounds are frequent in English and that their frequency follows a Zipfian distribution, with the majority of noun compounds encountered in text being rare types. 
This is consistent with the Romanian data: 
for most compounds, the annotators did not choose a labeled category but selected "none." Figure \ref{fig:freqs_category_heads_modifiers} shows the average frequency of the heads and modifiers (of the all the extracted noun compounds in the Romanian Universal Dependency  Treebank) in each category as a function of how many times that category was selected. We compute this in the following way: for a given category, we inspect all heads (modifiers) of compounds that were annotated with this category; for each head (modifier), we compute how often it appears in any compound in the corpus, not just the 1000 that were annotated; finally, we average this number over all heads (modifiers) of compounds annotated with this category. We did this for all 17 categories. 

Focusing on the compounds annotated with "none" of the labeled categories, we find that the each head of those compounds appears, on average, in 2.72 compounds in the corpus. For the modifier nouns in the "none" category, the average corpus frequency is 2.33. For the most frequently chosen category, "process + undergoer," the average frequency of the head and the modifier nouns in the corpus is 2.83 and 4.29, respectively. This suggests that the majority of noun compounds encountered in text are rare types in Romanian (as they are in English), which makes interpretation and and categorization challenging.\footnote{It is striking that the modifier nouns are far more frequent than the heads; currently we have no explanation.}

Among the compounds with "none" annotations, 340 were of type N Prep N and 149 were of type NN (with the second nouns inflected for genitive). Note that this was computed  checking the type of each compound that was labeled "none," so if a compound received the label "none" from both annotators, it was counted twice.

\textbf{Model predictions.} Our model's predictions agreed with 
the human annotations in 169 out of 250 (68\%) test compounds. Out of these, forty-six were classified identically by both annotators and the neural network.
The confusion matrices for the test set, for both human annotations (Table \ref{tab:confusion_annotators}) and our model's predictions (Table \ref{tab:confusion_nn}) show significant similarity. 
The cells represent how often the category with the ID = row number was chosen by one annotator and the category with ID = column number, by the other one. 
The diagonal cells represent the agreements for that category.

The "none" category was the most frequently chosen and the most frequently agreed-upon one, and some labeled categories were rarely  selected.

\begin{table}
    \centering
    \resizebox{\columnwidth}{!}{%
    \begin{tabular}{l|*{17}{c}}
                 Category ID&1&2&3&4&5&6&7&8&  9&  10&  11&  12&  13&  14&  15&  16& 17\\
                 \hline
                 1&\textbf{32}&\textbf{24}&\textbf{39}&0&0&\textbf{5}&\textbf{2}&\textbf{15}&  \textbf{3}&  \textbf{14}&  \textbf{9}&  \textbf{1}&  \textbf{5}&  0&  0&  0& 0\\
                 2&\textbf{24}&\textbf{22}&\textbf{11}&0&\textbf{1}&0&0&\textbf{3}&  0&  \textbf{6}&  0&  \textbf{6}&  0&  0&  0&  0& \textbf{2}\\
                 3&\textbf{39}&\textbf{11}&\textbf{24}&\textbf{1}&\textbf{3}&0&0&\textbf{7}&  0&  0&  0&  0&  \textbf{3}&  0&  0&  0& 0\\
                 4&0&0&\textbf{1}&\textbf{1}&0&0&0&0&  0&  0&  0&  0&  \textbf{1}&  0&  0&  0& 0\\
                 5&0&\textbf{1}&\textbf{3}&0& \textbf{1}&0&0&0&  0&  0&  0&  0&  0&  0&  0&  0& 0\\
                 6&\textbf{5}&0&0&0&0&0&0&0&  0&  0&  0&  0&  0&  0&  0&  0& 0\\
                 7&\textbf{2}&0&0&0&0&0&0&0&  0&  0&  0&  0&  0&  0&  0&  0& 0\\
                 8&\textbf{15}&\textbf{3}&\textbf{7}&0&0&0&0&\textbf{2}&  0&  0&  0&  \textbf{1}&  0&  0&  0&  0& 0\\
              9&\textbf{3}& 0& 0& 0& 0& 0& 0& 0& 0& 0& 0& 0& 0& 0& 0& 0&0\\
              10&\textbf{14}& \textbf{6}& 0& 0& 0& 0& 0& 0& 0& \textbf{1}& 0& \textbf{3}& 0& 0& 0& 0&0\\
              11&\textbf{9}& 0& 0& 0& 0& 0& 0& 0& 0& 0& \textbf{1}& 0& 0& 0& 0& 0&0\\
              12&\textbf{1}& \textbf{6}& 0& 0& 0& 0& 0& \textbf{1}& 0& \textbf{3}& 0& 0& 0& 0& 0& 0&0\\
              13&\textbf{5}& 0& \textbf{3}& \textbf{1}& 0& 0& 0& 0& 0& 0& 0& 0& \textbf{1}& 0& 0& 0&0\\
              14&0& 0& 0& 0& 0& 0& 0& 0& 0& 0& 0& 0& 0& 0& 0& 0&0\\
              15&0& 0& 0& 0& 0& 0& 0& 0& 0& 0& 0& 0& 0& 0& 0& 0&0\\
              16&0& 0& 0& 0& 0& 0& 0& 0& 0& 0& 0& 0& 0& 0& 0& 0&0\\
                 17&0&\textbf{2}&0&0&0&0&0&0&  0&  0&  0&  0&  0&  0&  0&  0& 0\\
    \end{tabular}
    }
    \caption{Confusion matrix for human annotations. Both the rows and the categories refer to categories identifiers. 
    }
    \label{tab:confusion_annotators}
\end{table}

\begin{table}
    \resizebox{\columnwidth}{!}{%
    \begin{tabular}{l|*{17}{c}}
                 Category ID&1&2&3&4&5&6&7&8&  9&  10&  11&  12&  13&  14&  15&  16& 17\\
                 \hline
                 1&\textbf{28}&\textbf{9}&\textbf{11}&\textbf{4}&\textbf{1}&\textbf{4}&\textbf{2}&\textbf{10}&  \textbf{2}&  \textbf{4}&  \textbf{6}&  \textbf{5}&  \textbf{4}&  \textbf{1}&  \textbf{7}&  0& \textbf{1}\\
                 2&\textbf{9}&\textbf{20}&\textbf{7}&\textbf{1}&\textbf{6}&0&0&\textbf{1}&  0&  \textbf{8}&  \textbf{2}&  0&  0&  0&  \textbf{2}&  0& \textbf{7}\\
                 3&\textbf{11}&\textbf{7}&\textbf{12}&\textbf{2}&\textbf{1}&\textbf{2}&0&  \textbf{4}& \textbf{1}&  \textbf{4}&  \textbf{1}&  0&  \textbf{1}&  0&  0&  \textbf{1}& 0\\
                 4&\textbf{4}&\textbf{1}&\textbf{2}&\textbf{5}&0&0&0&\textbf{2}&  0&  0&  \textbf{2}&  \textbf{2}&  \textbf{3}&  0&  \textbf{1}&  0& 0\\
                 5&\textbf{1}&\textbf{6}&\textbf{1}&0&\textbf{4}&0&0&\textbf{1}&  0&  \textbf{1}&  0&  \textbf{1}&  0&  0&  \textbf{1}&  \textbf{1}& 0\\
                 6&\textbf{4}&0&\textbf{2}&0&0&\textbf{1}&0&\textbf{1}&  0&  0&  0&  0&  0&  \textbf{2}&  0&  0& 0\\
                 7&\textbf{2}&0&0&0&0&0&0&0&  0&  0&  0&  0&  0&  0&  0&  0& 0\\
                 8&\textbf{10}&\textbf{1}&\textbf{4}&\textbf{2}&\textbf{1}&\textbf{1}&0&\textbf{2}&  0&  \textbf{7}&  0&  \textbf{4}&  \textbf{2}&  0&  0&  0& \textbf{1}\\
              9 &\textbf{2}& 0& 0& \textbf{1}& 0& 0& 0& 0& \textbf{2}& 0& 0& 0& \textbf{2}& 0& 0& 0&0\\
              10&\textbf{4}& \textbf{8}& \textbf{8}& \textbf{4}& \textbf{1}& 0& 0& \textbf{7}& 0& \textbf{3}& 0& \textbf{1}& 0& \textbf{2}& 0& 0&\textbf{1}\\
              11&\textbf{6}& \textbf{2}& \textbf{2}& \textbf{1}& 0& 0& 0& 0& 0& 0& 0& 0& 0& 0& \textbf{1}& 0&0\\
              12&\textbf{5}& 0& 0& 0& \textbf{1}& 0& 0& \textbf{4}& 0& \textbf{1}& 0& \textbf{1}& 0& 0& 0& \textbf{1}&\textbf{2}\\
              13&\textbf{4}& 0& 0& \textbf{1}& 0& 0& 0& \textbf{2}& \textbf{2}& 0& 0& 0& \textbf{5}& 0& 0& 0&0\\
              14&\textbf{1}& 0& 0& 0& 0& \textbf{2}& 0& 0& 0& \textbf{2}& 0& 0& 0& \textbf{1}& 0& 0&0\\
              15&\textbf{7}& \textbf{2}& \textbf{2}& 0& \textbf{1}& 0& 0& 0& 0& 0& \textbf{1}& 0& 0& 0& 0& \textbf{1}&0\\
              16&0& 0& 0& \textbf{1}& \textbf{1}& 0& 0& 0& 0& 0& 0& \textbf{1}& 0& 0& \textbf{1}& \textbf{1}&0\\
              17&\textbf{1}& \textbf{7}& \textbf{7}& 0& 0& 0& 0& \textbf{1}& 0& \textbf{1}& 0& \textbf{2}& 0& 0& 0& 0& 0\\
    \end{tabular}
    }
    \caption{Confusion matrix for the neural network. For each compound we consider the two labels with highest probability predicted by the neural network.}
    \label{tab:confusion_nn}
\end{table}

\section{Discussion}

Interpreting noun compounds poses a necessary and significant challenge to  Natural Language Processing, as noun compounds are frequent, highly productive and their meanings compose in multiple, subtle ways. A key difficulty lies in identifying a comprehensive set of meaning relations between the constituent members that reflects speakers construal when processing the compounds. 
The low agreement rate for labeled relations that we found among the human annotators  indicates that the proposed taxonomy is insufficient either in number or type for capturing human semantic interpretation.  An intriguing finding was that the "none" category was most frequently selected and agreed upon by the human annotators, and that most compounds were also classified as "none" by the neural network. This, too, indicates that our relation inventory is insufficient, at least for the Romanian compounds. 

We had stripped the prepositions from the NPN compounds that were input to the neural network so that the dimensionality could be preserved regardless of compound syntax. The model's predictions were positively correlated with the human predictions, suggesting that the preposition does not bear significantly on the meaning of the compound. Similarly, no measurable effect was found for compounds with a case inflection on the second constituent.

\section{Conclusion}
Noun compound interpretation, while effortlessly performed by humans, continues to pose a challenge to automatic systems. We hope that future research will converge on an inventory of semantic categories that both humans and machines can discriminate and interpret well enough for applications like translation and question answering. Our work invites comparative investigations in other languages, particularly understudied languages and languages where morphosyntax might provide clues to the semantic interpretation. 

\section*{Ethics statement}
There are no ethical concerns about this work.

\section*{Conflict of interest}
We declare no conflict of interest. 

\section*{Data availability}
The Romanian compounds and all judgments are available here: \url{https://anonymous.4open.science/r/NounCompounds-4108/}

\nocite{*}
\section{Bibliographical References}\label{sec:reference}

\bibliographystyle{lrec-coling2024-natbib}
\bibliography{lrec-coling2024-example}


\end{document}